\documentclass[runningheads]{llncs}
\usepackage[T1]{fontenc}
\usepackage{graphicx}
\usepackage{booktabs,tabularx}
\usepackage{xcolor}
\usepackage[table]{xcolor} 
\usepackage{amsmath}
\usepackage{amssymb}
\usepackage{multirow}
\newcommand{\mstd}[2]{\begin{tabular}{c} #1 \\ {\scriptsize $\pm$ #2} \end{tabular}}

\begin{document}
\title{Self-Supervised Learning of Plant Image Representations}
%
%

\author{Ilyass Moummad\inst{1}\orcidID{0009-0003-2925-2500} \and
Kawtar Zaher\inst{1,2}\orcidID{0009-0008-9920-9483} \and
Hervé Goëau\inst{3}\orcidID{0000-0003-3296-3795} \and
Jean-Christophe Lombardo\inst{1}\orcidID{0000-0002-9656-4219} \and
Pierre Bonnet\inst{3}\orcidID{0000-0002-2828-4389} \and
Alexis Joly\inst{1}\orcidID{0000-0002-2161-9940}}
\authorrunning{I. Moummad et al.}
%
\institute{INRIA, LIRMM, Université de Montpellier, Montpellier, France \and
Institut National de l’Audiovisuel, Paris, France
\and
AMAP, Université de Montpellier, CIRAD, INRAE, IRD, Montpellier, France\\
\email{\{ilyass.moummad,alexis.joly\}@inria.fr}}
\maketitle              
\begin{abstract}
Automated plant recognition plays a crucial role in biodiversity monitoring and conservation, yet current approaches rely heavily on supervised learning, which is limited by the availability of expert-labeled data. Self-supervised learning (SSL) offers a scalable alternative, but existing methods and training protocols are largely designed for coarse-grained visual tasks and may not transfer well to fine-grained domains such as plant species recognition. In this work, we investigate SSL for plant image representation learning. We show that commonly used augmentations in SSL pipelines—such as Gaussian blur, grayscale conversion, and solarization—are detrimental in the context of plant images, as they remove subtle discriminative cues essential for fine-grained recognition. We instead identify alternative transformations, including affine and posterization, that are better suited to this domain. We further demonstrate that training SimDINOv2 on the iNaturalist 2021 Plantae subset yields significantly stronger representations than training on ImageNet-1K, highlighting the importance of domain-specific data for SSL. Our findings are consistent across both ViT-Base and ViT-Large architectures. Moreover, our models achieve competitive performance and sometimes outperform strong supervised baselines Pl@ntCLEF and BioCLIP, on downstream plant recognition tasks in few-shot settings. Overall, our results highlight the critical importance of domain-adapted augmentation strategies and dataset selection in self-supervised learning, and provide practical guidelines for building scalable models for biodiversity monitoring.\footnote{Code and models: \url{https://github.com/ilyassmoummad/sslplant}}
\end{abstract}

\section{Introduction}

Automatic plant recognition plays a crucial role in biodiversity monitoring, ecological research, and precision agriculture~\cite{plantmonitor}. Large scale platforms such as Pl@ntNet~\cite{plantnetapp} have demonstrated the potential of computer vision systems to assist both experts and citizens in identifying plant species in the wild. However, these systems rely heavily on supervised learning, which requires large amounts of labeled data. In the context of plant species recognition, acquiring such annotations is particularly challenging, as it demands domain expertise and significant manual effort~\cite{plantclef2023}. This limitation raises important concerns regarding scalability and coverage, especially for under-documented ecosystems.

Self-supervised learning (SSL)~\cite{sslcookbook} offers a promising alternative by enabling representation learning without manual annotations. While SSL has achieved remarkable success in general-purpose visual representation learning~\cite{moco,simclr,swav,byol,simsiam,barlow,dino,dinov2,simdino,bits}, its application to plant image analysis remains relatively underexplored. Most state-of-the-art SSL methods are designed and evaluated on coarse-grained datasets such as ImageNet~\cite{imagenet}, where categories differ significantly (e.g., distinguishing between animals and objects). In this setting, invariance to strong data augmentations such as aggressive color transformations, blurring, or solarization—has proven beneficial~\cite{barlow,dino}. However, plant species recognition is inherently a fine-grained task, where subtle visual cues such as texture, venation patterns, and color distributions are critical for discrimination. As a result, augmentation strategies that are effective for coarse-grained tasks may be detrimental in this domain. For example, color-based transformations like solarization or grayscale conversion can distort or remove key discriminative features, as illustrated in Figure~\ref{fig:aug}. This highlights the need for domain-adapted augmentation policies in SSL for biodiversity applications.

In this work, we investigate the recently proposed SimDINOv2~\cite{simdino} framework for self-supervised representation learning on plant images. SimDINOv2 builds upon DINOv2~\cite{dinov2} while simplifying its training pipeline by removing several heuristic components and introducing an explicit coding rate regularization to prevent representation collapse. We conduct a systematic ablation study of commonly used augmentations and demonstrate that widely adopted transformations such as Gaussian blur, grayscale conversion, and solarization negatively impact performance in the context of plant recognition.

Furthermore, we introduce and evaluate two alternative transformations, posterization and affine transformations, which have been largely overlooked in SSL pipelines~\cite{dassl}. Our results show that these augmentations are better suited to preserving fine-grained visual characteristics of plant species, leading to improved representation quality. This work emphasizes the importance of domain-specific design choices in self-supervised learning and provides practical insights for developing more effective models for biodiversity monitoring.

\section{Related Work}

\textbf{Automatic plant recognition} plays a central role in biodiversity monitoring~\cite{plantmonitor} and has been significantly advanced by citizen science platforms that collect large-scale visual observations~\cite{plantnetapp}. Applications such as ObsIdentify~\cite{obsidentify}, iNaturalist~\cite{inaturalist}, Flora Incognita~\cite{floraincognita}, and Pl@ntNet~\cite{plantnetapp} allow users to identify plant species from images while continuously enriching training datasets through user contributions. Among these, Pl@ntNet~\cite{plantnetapp} stands out as a large-scale system dedicated to plant identification, covering over 80,000 species and supporting fine-grained recognition, including intra-species variability as well as plant diseases and pests. Despite these advances, most existing systems rely on supervised learning, which limits scalability due to the high cost and expertise required for annotation.

\textbf{Large-scale plant datasets} have been introduced to support plant recognition, covering both species identification and plant pathology. Pl@ntNet-300K~\cite{plantnet300k} contains approximately 300,000 images across 1,000 species and is widely used as a benchmark for supervised classification. Larger collections are available through citizen science initiatives, notably the iNaturalist 2021 dataset~\cite{inaturalist2021}, whose Plantae subset comprises around 1.1 million images spanning more than 4,000 species. Its scale and diversity, together with its similarity to ImageNet-1K, make it particularly well suited for training data-intensive self-supervised models. Other datasets, such as Deep-Plant-Disease~\cite{deepplantdisease}, focus on plant pathology and provide labeled examples of diseases across multiple crops. In this work, we leverage the iNaturalist 2021 Plantae subset due to its scale and relevance for self-supervised pretraining.

\textbf{Self-supervised learning} has emerged as a powerful paradigm for learning visual representations without manual annotations. A dominant approach is instance discrimination, where models are trained to produce invariant representations across augmented views of the same image. Early methods rely on contrastive objectives with explicit negative samples (e.g., SimCLR~\cite{simclr}, MoCo~\cite{moco}), while clustering-based approaches such as SwAV~\cite{swav} improve scalability. More recent methods eliminate the need for negatives and instead prevent representation collapse through implicit regularization, including asymmetric self-distillation (BYOL~\cite{byol}, SimSiam~\cite{simsiam}, DINO~\cite{dino}) and redundancy reduction objectives (Barlow Twins~\cite{barlow}, VICReg~\cite{vicreg}).

These approaches primarily focus on learning global representations. DINOv2~\cite{dinov2} extends this paradigm by incorporating a patch-level objective, enabling the model to capture both global semantics and local structures. Building on this, SimDINOv2~\cite{simdino} simplifies DINOv2~\cite{dinov2} by removing heuristic mechanisms such as centering and sharpening, and instead enforces representation diversity through coding-rate regularization~\cite{codingrate}, while relying on a simple cosine similarity objective for alignment.

Despite their strong performance on generic benchmarks, SSL methods rely heavily on augmentation invariance designed for coarse-grained recognition. Their application to fine-grained domains such as plant species recognition remains underexplored, particularly with respect to augmentation design, which can critically impact the preservation of subtle discriminative features.

\section{PlantSSL: Self-Supervised Representation Learning of Plant Images}

\subsection{Preliminaries: Contrastive Learning}

Contrastive learning aims to learn representations that are invariant across different augmented views of the same image while remaining discriminative across different images.

\paragraph{SimCLR.}
Given an input image $x$, a set of augmented views $\mathcal{V}(x)$ is generated using a stochastic augmentation pipeline. Let $z(v) \in \mathbb{R}^d$ denote the $\ell_2$-normalized representation of view $v$, obtained from an encoder $f_\theta$.

Let $\mathcal{B}$ be a batch of images, and define the set of all views in the batch as
\[
\mathcal{V}_{\mathcal{B}} = \bigcup_{x \in \mathcal{B}} \mathcal{V}(x).
\]
For a given anchor view $v \in \mathcal{V}(x)$, we denote by $v^+$ another view of the same image $x$, and by
\(
\mathcal{A}(v) = \mathcal{V}_{\mathcal{B}} \setminus \{v\}
\)
the set of all other views in the batch.

The SimCLR loss (also known as InfoNCE or NT-Xent) is defined as:
\begin{equation}
\mathcal{L}_{\text{SimCLR}} =
\frac{1}{|\mathcal{V}_{\mathcal{B}}|} 
\sum_{v \in \mathcal{V}_{\mathcal{B}}}
- \log 
\frac{
\exp\left( \langle z(v), z(v^+) \rangle / \tau \right)
}{
\sum\limits_{v' \in \mathcal{A}(v)}
\exp\left( \langle z(v), z(v') \rangle / \tau \right)
},
\end{equation}
where $\tau$ is a temperature parameter. This objective pulls together representations of different views of the same image while pushing apart representations from other images in the batch.

\paragraph{SupCon.}
Supervised Contrastive Learning extends SimCLR by leveraging label information to define multiple positive pairs.

Let $y(x)$ denote the label of image $x$. For an anchor view $v \in \mathcal{V}(x)$, we define the set of positives as
\(
\mathcal{P}(v) = \{ v' \in \mathcal{V}_{\mathcal{B}} \setminus \{v\} \;:\; y(v') = y(x) \},
\)
i.e., all views in the batch whose underlying images share the same label.

The SupCon loss is then given by:
\begin{equation}
\mathcal{L}_{\text{SupCon}} =
\frac{1}{|\mathcal{V}_{\mathcal{B}}|} 
\sum_{v \in \mathcal{V}_{\mathcal{B}}}
\frac{-1}{|\mathcal{P}(v)|}
\sum_{v^+ \in \mathcal{P}(v)}
\log 
\frac{
\exp\left( \langle z(v), z(v^+) \rangle / \tau \right)
}{
\sum\limits_{v' \in \mathcal{A}(v)}
\exp\left( \langle z(v), z(v') \rangle / \tau \right)
}.
\end{equation}

This formulation encourages representations of samples from the same class to cluster together, providing an upper bound on the performance achievable by purely self-supervised methods.

\paragraph{Role of data augmentation.} The definition of positive pairs, and therefore the invariances learned by the model, is entirely determined by the augmentation pipeline $\mathcal{A}$. As a result, the choice of augmentations plays a critical role in shaping the learned representation, particularly in fine-grained settings where subtle visual cues must be preserved.

\subsection{Data Augmentation for Plant Images}

\begin{figure}[h]
  \centering
  \includegraphics[width=1.\columnwidth]{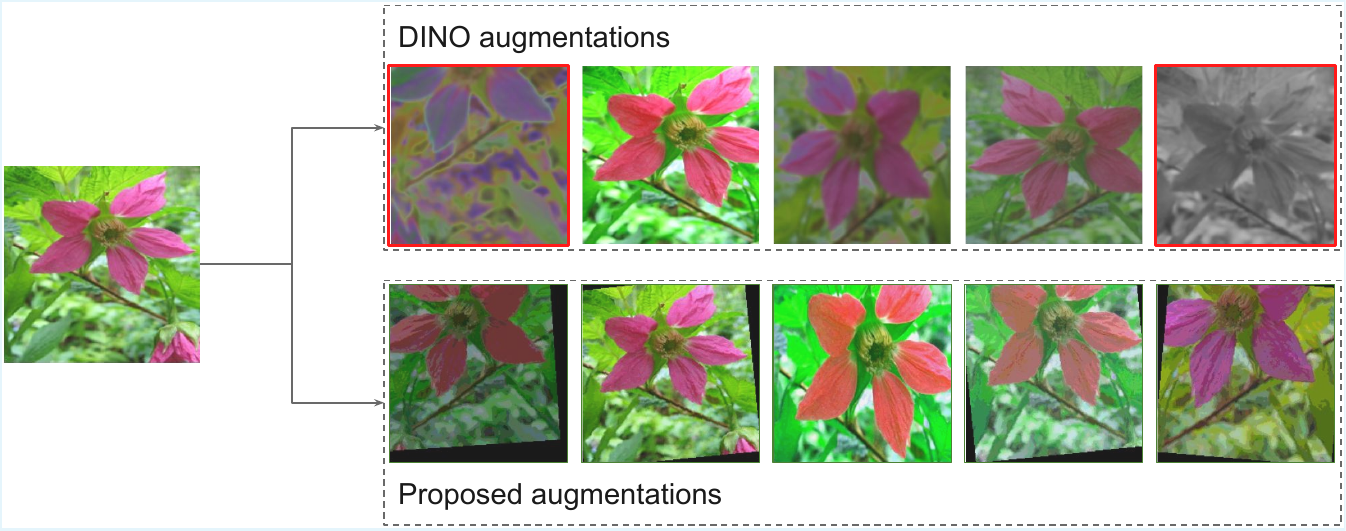}
    \caption{Comparison of general-purpose data augmentations from self-supervised visual learning frameworks (e.g., SimCLR, BYOL, DINO) with our proposed augmentations for fine-grained plant species recognition. Standard techniques such as Gaussian blur, grayscale, and polarize destroy crucial plant information, whereas posterize and affine transformations preserve distinguishing features while still allowing visual diversity.}
  \label{fig:aug}
\end{figure}

Data augmentation plays a central role in SSL~\cite{sslcookbook}, as the learning objective enforces invariance between different views of the same image. Standard augmentation pipelines, popularized by methods such as SimCLR~\cite{simclr}, BYOL~\cite{byol}, and DINO~\cite{dino}, include strong appearance transformations such as color jittering, grayscale conversion, Gaussian blur, and solarization. While effective for coarse-grained recognition, these transformations can significantly alter or remove subtle visual cues—such as color distributions, textures, and venation patterns—that are essential for plant species discrimination (Figure~\ref{fig:aug}, top row).

To address this limitation, we propose a plant-adapted augmentation strategy that preserves fine-grained characteristics while still generating sufficiently diverse views. Specifically, we replace grayscale conversion, Gaussian blur, and solarization with two alternative transformations: \textit{posterization} and \textit{affine transformations}. Posterization reduces the number of color levels, producing appearance variations while retaining global color structure. Affine transformations (including rotation, translation, scaling, and shearing) introduce geometric variability without degrading local textures. As illustrated in Figure~\ref{fig:aug} (bottom row), these transformations maintain discriminative plant features while increasing view diversity.

\paragraph{Experimental validation.}
To isolate the impact of data augmentation independently of architectural and optimization choices, we conduct a controlled study using contrastive learning (SimCLR) and its supervised counterpart (SupCon). This setup allows us to evaluate how augmentation choices affect representation quality relative to label supervision.

All models are trained from scratch using a ResNet-18 backbone on a subset of the iNaturalist 2021 Plantae dataset, comprising 213,550 training images across 4,271 species. We use random resized crop and horizontal flip as base augmentations, denoted $\mathcal{A}_{\text{base}}$, to ensure sufficient variability and avoid collapsed representations. Models are trained for 100 epochs, and the learned representations are evaluated using $k$-nearest neighbors classification on the validation subset containing 42,710 images (10 images per species).

\begin{table}[h]
\centering
\caption{Impact of data augmentations for self-supervised learning of plant image representations on the iNaturalist 2021 Plantae subset (4,271 species). 
We train ResNet18 from scratch on training mini subset (213,550 images) for 100 epochs using contrastive learning and evaluate on the validation subset (42,710 images, 10 images per species). $\mathcal{A}_{\text{base}}$ denotes Random Resized Crop and Horizontal Flip.} 

\label{tab:aug}
\begin{tabular}{l l c}
\toprule
\textbf{Method} & \textbf{Augmentations} & \textbf{Acc. (\%)} \\
\midrule

\multicolumn{3}{l}{\textit{Reference}} \\
SupCon (w/ labels) & $\mathcal{A}_{\text{base}}$ & 10.61 \\
SimCLR & $\mathcal{A}_{\text{base}}$ & 3.70 \\
\midrule

\multicolumn{3}{l}{\textit{DINO augmentations}} \\
SimCLR & $\mathcal{A}_{\text{base}}$ + Jitter + Grayscale + Blur + \textcolor{red}{Solarize} & 9.70 \\
SimCLR & $\mathcal{A}_{\text{base}}$ + Jitter + Grayscale + Blur & 9.96 \\


\multicolumn{3}{l}{\textit{Plant-adapted augmentations (ours)}} \\
SimCLR & $\mathcal{A}_{\text{base}}$ + Jitter + Posterize + Affine & \textbf{10.35} \\
SimCLR & $\mathcal{A}_{\text{base}}$ + Jitter + Posterize + Affine + \textcolor{red}{Grayscale} & 9.83 \\
SimCLR & $\mathcal{A}_{\text{base}}$ + Jitter + Posterize + Affine + \textcolor{red}{Blur} & 8.76 \\
\bottomrule
\end{tabular}
\end{table}

Results are reported in Table~\ref{tab:aug}. Using only $\mathcal{A}_{\text{base}}$, SimCLR performs poorly compared to its supervised counterpart, highlighting the difficulty of learning fine-grained discriminative features for plant species recognition. Incorporating standard augmentations from prior SSL methods (e.g. DINO) significantly improves performance; however, removing solarization yields further gains, suggesting that this transformation is detrimental in the plant domain. Our proposed augmentation strategy achieves the best performance, nearly matching the supervised SupCon upper bound. Importantly, ablation results show that reintroducing grayscale or Gaussian blur consistently degrades performance. These findings confirm that commonly used SSL augmentations are not well suited for fine-grained plant recognition, and that carefully designed, domain-adapted transformations are critical for learning discriminative representations.

These observations motivate the use of plant-adapted augmentations in modern SSL frameworks, which we explore next.

\subsection{SimDINOv2 for Plant Representation Learning}

Building on the importance of augmentation design highlighted in the previous section, we integrate plant-adapted augmentations into a modern multi-view self-supervised framework. We adopt SimDINOv2, which combines self-distillation with explicit regularization to prevent representation collapse, while jointly learning global and local features.

\paragraph{Teacher--student framework.}
Given an input image $x$, we generate a set of augmented views. We denote by $\mathcal{V}_g$ the set of global views and by $\mathcal{V}_l$ the set of local views. The teacher network $f_{\theta_t}$ processes only global views $v \in \mathcal{V}_g$, while the student network $f_{\theta_s}$ processes all views $v \in \mathcal{V}_g \cup \mathcal{V}_l$.

For a given view $v$, a Vision Transformer produces a global CLS token representation denoted by $z^{\text{cls}}(v)$, and a set of patch-level token representations denoted by $\{ z^i(v) \}_{i=1}^{N}$. All representations are $\ell_2$-normalized, so that dot products correspond to cosine similarity.

\textit{Global (CLS) alignment.}
The student is trained to match the teacher's CLS representations from global views across all student views:
\begin{equation}
\mathcal{L}_{\text{cls}} =
\frac{1}{|\mathcal{B}|} \sum_{x \in \mathcal{B}}
\frac{1}{|\mathcal{V}_g| \, |\mathcal{V}_g \cup \mathcal{V}_l|}
\sum_{v_t \in \mathcal{V}_g} \sum_{v_s \in \mathcal{V}_g \cup \mathcal{V}_l}
\left( 1 - \langle z^{\text{cls}}_s(v_s), z^{\text{cls}}_t(v_t) \rangle \right).
\end{equation}

\textit{Patch-level alignment.}
In addition, a masked patch prediction objective inspired by iBOT is applied to patch tokens. For each image, a global view $v \in \mathcal{V}_g$ is processed by both networks. The teacher receives the full (unmasked) sequence of patch tokens, while the student operates on a masked version of the same view, where a subset of patch tokens is replaced by mask tokens after patch embedding.

Let $\mathcal{M}(v) \subset \{1, \dots, N\}$ denote the set of masked patch indices. The student is trained to match the teacher representations at these locations:
\begin{equation}
\mathcal{L}_{\text{patch}} =
\frac{1}{|\mathcal{B}|} \sum_{x \in \mathcal{B}}
\frac{1}{|\mathcal{M}(v)|}
\sum_{i \in \mathcal{M}(v)}
\left( 1 - \langle z^i_s(v), z^i_t(v) \rangle \right).
\end{equation}

This objective encourages the model to infer missing local information from context while producing consistent and semantically meaningful patch-level representations.

\paragraph{Coding rate regularization.}
To further prevent representation collapse, SimDINOv2 incorporates a coding rate regularization term applied to the student CLS representations. Let $Z_s \in \mathbb{R}^{B \times d}$ denote the matrix of $\ell_2$-normalized CLS features for a batch of size $B$. The regularization term is defined as:
\begin{equation}
\mathcal{L}_{\text{reg}} = - \frac{1}{2} \log \det \left( I + \frac{d}{B \varepsilon} Z_s^\top Z_s \right),
\end{equation}
where $\varepsilon > 0$ controls the precision of the representation in the rate--distortion sense. This term encourages the representations to span a large volume in feature space, promoting diversity and preventing collapse.

\paragraph{Final objective.}
The overall training objective is a weighted combination of the different terms:
\begin{equation}
\mathcal{L} = 
\lambda_{\text{cls}} \mathcal{L}_{\text{cls}} 
+ \lambda_{\text{patch}} \mathcal{L}_{\text{patch}} 
+ \lambda_{\text{reg}} \mathcal{L}_{\text{reg}}.
\end{equation}

The teacher parameters are updated as an exponential moving average (EMA) of the student:
\begin{equation}
\theta_t \leftarrow \lambda \theta_t + (1 - \lambda)\theta_s.
\end{equation}

\begin{figure}[h]
  \centering
  \includegraphics[width=.9\columnwidth]{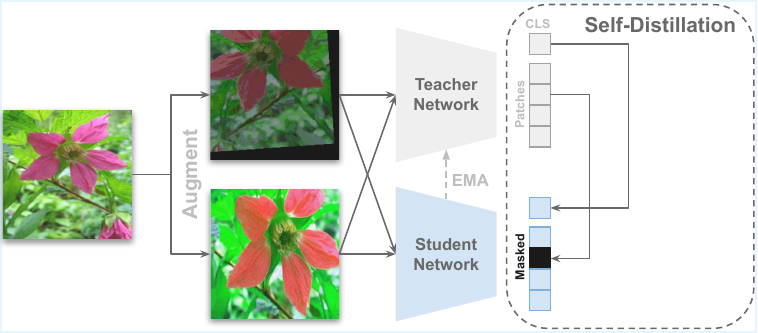}
    \caption{Self-supervised learning of plant image representations using SimDINOv2 framework. An input image is augmented to produce two views: one passes through a teacher network to generate target representations, while the other passes through a student network that learns to match the teacher. Both global CLS token and patch-level representations are trained via self-distillation with cosine similarity. The student is optimized with backpropagation, and the teacher is updated using an exponential moving average of the student. The coding rate regularization is applied to the student model to prevent collapsed representations.}
  \label{fig:ssl}
\end{figure}

Figure~\ref{fig:ssl} illustrates SimDINOv2 framework, showing both global and patch-level self-distillation using cosine similarity.

\section{Experiments}

We evaluate the effectiveness of our approach for self-supervised learning of plant image representations through few-shot classification on diverse plant benchmarks. Our experiments aim to answer two main questions: (i) whether domain-specific pretraining on plant data improves representation quality compared to generic datasets such as ImageNet, and (ii) whether our approach can compete with strong supervised baselines.

\subsection{Experimental Setup}

\paragraph{Pretraining.}
We consider two sources of pretrained models. First, we use publicly available SimDINOv2 models trained on ImageNet-1K. Second, we train SimDINOv2 from scratch on the iNaturalist 2021 Plantae subset, which contains approximately 1.1M images across 4,271 plant species and is comparable in scale to ImageNet. This setting allows us to isolate the impact of domain-specific data. We evaluate both ViT-Base (B) and ViT-Large (L) architectures.

\paragraph{Evaluation protocol.}
We adopt the few-shot evaluation protocol introduced in BioCLIP using the SimpleShot~\cite{simpleshot} classifier. Given a small support set (1-shot or 5-shot), class prototypes are computed as the mean of L2-normalized feature embeddings, and query samples are classified via nearest-neighbor search in the embedding space.

\paragraph{Datasets.}
While BioCLIP evaluates on a diverse set of biological domains, we restrict our study to plant-related datasets from the MetaAlbum benchmark. Specifically, we consider Pl@ntNet, PlantVillage, Medicinal Leaf, and PlantDoc, covering both plant species recognition and plant disease classification tasks. A summary of the datasets is provided in Tab.~\ref{tab:datasets}.

\begin{table}[t]
\centering
\caption{Summary of evaluation datasets from MetaAlbum.}
\resizebox{\columnwidth}{!}{
\begin{tabular}{l l c c}
\hline
\textbf{Dataset} & \textbf{Description} & \textbf{\#Samples} & \textbf{\#Classes} \\
\hline
PlantNet & Citizen science species-labeled plant images & 1{,}000 & 25 \\
PlantVillage & Museum-style leaf specimens & 1{,}520 & 38 \\
Med.\ Leaf & Leaves from mature healthy medicinal plants & 1{,}040 & 26 \\
PlantDoc & 17 diseases across 13 plant species & 1{,}080 & 27 \\
\hline
\end{tabular}
}
\label{tab:datasets}
\end{table}

\paragraph{Baselines.}
We compare our approach against strong baselines commonly used for plant recognition. Pl@ntCLEF~\cite{plantclef2024} is a DINOv2~\cite{dinov2} model fine-tuned in a supervised manner on the Pl@ntCLEF-2024 dataset, which represents the flora of southwestern Europe with 1.4M images covering 7,800 plant species. BioCLIP~\cite{bioclip} is a vision–language model based on CLIP~\cite{clip}, adapted to biodiversity data and fine-tuned on large-scale image–text pairs from TreeOfLife-10M. On the self-supervised side, we consider SimDINOv2~\cite{simdino} models pretrained on ImageNet-1K as a reference, allowing us to isolate the impact of domain-specific pretraining. Altogether, these baselines provide a comprehensive comparison across supervised, multimodal, and self-supervised paradigms. Table~\ref{tab:fewshot} reports few-shot classification accuracy.

\paragraph{Implementation details.}
For self-supervised training, we closely follow the SimDINOv2 training protocol~\cite{simdino}. We train both ViT-Base and ViT-Large models using the AdamW optimizer, adopting the same optimization hyperparameters, including the learning rate schedule and warm-up strategy. We employ the same multi-crop augmentation scheme consisting of 2 global views with resolution $224 \times 224$ and 10 local views with resolution $96 \times 96$. 

\subsection{Few-Shot Classification Results}

\paragraph{Impact of domain-specific pretraining.}
Training SimDINOv2 on iNaturalist Plantae significantly improves performance over ImageNet-pretrained models across all datasets and shot settings. For example, on Pl@ntNet, ViT-L improves from $35.6\%$ to $77.4\%$ in the 1-shot setting. Similar gains are observed across all benchmarks, highlighting the importance of domain-specific data for fine-grained plant recognition.

\paragraph{Comparison with supervised baselines.}
Self-supervised models trained on iNaturalist achieve competitive performance and often surpass supervised baselines such as Pl@ntCLEF and BioCLIP. In particular, the ViT-L model achieves state-of-the-art results on Pl@ntNet (1-shot) and PlantVillage (both 1-shot and 5-shot), and consistently outperforms BioCLIP on the more challenging PlantDoc dataset, which contains plant disease images unseen during training.

\paragraph{Effect of model scale.}
Scaling from ViT-Base to ViT-Large yields consistent performance improvements. However, these gains remain smaller than those obtained by switching from generic to domain-specific pretraining, indicating that dataset relevance is a more critical factor than model capacity in this setting. At this scale, the ViT-L model would likely benefit further from additional training data, suggesting that its capacity is not yet fully utilized.

\begin{table}[h]
\centering
\caption{Few-shot classification accuracy (mean $\pm$ std) on plant image datasets from MetaAlbum. PC and TOL denote the Pl@ntCLEF and TreeOfLife datasets.}
\label{tab:fewshot}
\small
\setlength{\tabcolsep}{4pt}

\resizebox{\columnwidth}{!}{
\begin{tabular}{l c l | cc | cc | cc | cc}
\toprule

& & & \multicolumn{2}{c}{Pl@ntNet} 
  & \multicolumn{2}{c}{PlantVillage}
  & \multicolumn{2}{c}{Med. Leaf}
  & \multicolumn{2}{c}{PlantDoc} \\
\cmidrule(lr){4-5} \cmidrule(lr){6-7} \cmidrule(lr){8-9} \cmidrule(lr){10-11}

\textbf{Model} & \textbf{Arch} & \textbf{Pretraining} 
& 1-shot & 5-shot & 1-shot & 5-shot & 1-shot & 5-shot & 1-shot & 5-shot \\

\midrule
\multicolumn{11}{l}{\textit{\textcolor{black!70}{Supervised Models}}} \\
\midrule

\textcolor{black!70}{Pl@ntCLEF} & \textcolor{black!70}{B} & \textcolor{black!70}{LVD-142M} $\rightarrow$ \textcolor{black!70}{PC'24} 
& \textcolor{black!70}{\mstd{72.9}{1.7}} & \textcolor{black!70}{\mstd{91.2}{1.3}}
& \textcolor{black!70}{\mstd{67.1}{2.0}} & \textcolor{black!70}{\mstd{85.6}{1.4}}
& \textcolor{black!70}{\mstd{93.6}{0.8}} & \textcolor{black!70}{\mstd{99.4}{0.3}}
& \textcolor{black!70}{\mstd{38.8}{1.2}} & \textcolor{black!70}{\mstd{52.6}{1.0}} \\

\textcolor{black!70}{BioCLIP} & \textcolor{black!70}{B} & \textcolor{black!70}{OpenAI} $\rightarrow$ \textcolor{black!70}{TOL-10M} 
& \textcolor{black!70}{\mstd{61.2}{2.0}} & \textcolor{black!70}{\mstd{83.3}{1.5}}
& \textcolor{black!70}{\mstd{57.6}{2.2}} & \textcolor{black!70}{\mstd{79.6}{0.8}}
& \textcolor{black!70}{\mstd{81.8}{1.2}} & \textcolor{black!70}{\mstd{95.5}{1.3}}
& \textcolor{black!70}{\mstd{26.6}{2.2}} & \textcolor{black!70}{\mstd{42.3}{1.2}} \\

\midrule
\multicolumn{11}{l}{\textit{Self-Supervised Models}} \\
\midrule

SimDINOv2 & B & ImageNet-1K
& \mstd{32.8}{2.9} & \mstd{54.1}{0.7}
& \mstd{55.1}{1.4} & \mstd{79.2}{1.0}
& \mstd{77.6}{2.4} & \mstd{93.1}{1.0}
& \mstd{23.9}{2.0} & \mstd{37.7}{1.6} \\

SimDINOv2 & L & ImageNet-1K
& \mstd{35.6}{3.2} & \mstd{56.5}{0.8}
& \mstd{57.3}{1.4} & \mstd{80.1}{1.5}
& \mstd{76.1}{2.7} & \mstd{93.9}{0.7}
& \mstd{23.5}{1.5} & \mstd{38.0}{1.6} \\


\underline{\textit{Ours}} & & & & & & & & & & \\

SimDINOv2 & B & iNat21 Plantae
& \mstd{\underline{73.4}}{4.4} & \mstd{\underline{87.9}}{0.7}
& \mstd{\underline{64.7}}{1.9} & \mstd{\underline{85.4}}{0.5}
& \mstd{\underline{85.9}}{1.6} & \mstd{\underline{97.8}}{0.7}
& \mstd{\textbf{31.5}}{1.3} & \mstd{\underline{47.8}}{1.3} \\

SimDINOv2 & L & iNat21 Plantae
& \mstd{\textbf{77.4}}{3.3} & \mstd{\textbf{88.2}}{1.6}
& \mstd{\textbf{68.0}}{1.9} & \mstd{\textbf{86.6}}{0.1}
& \mstd{\textbf{87.8}}{1.7} & \mstd{\textbf{98.5}}{0.5}
& \mstd{\underline{31.2}}{2.0} & \mstd{\textbf{49.0}}{1.1} \\

\bottomrule
\end{tabular}
}
\end{table}

\begin{figure}[h]
  \centering
  \includegraphics[width=1.\columnwidth]{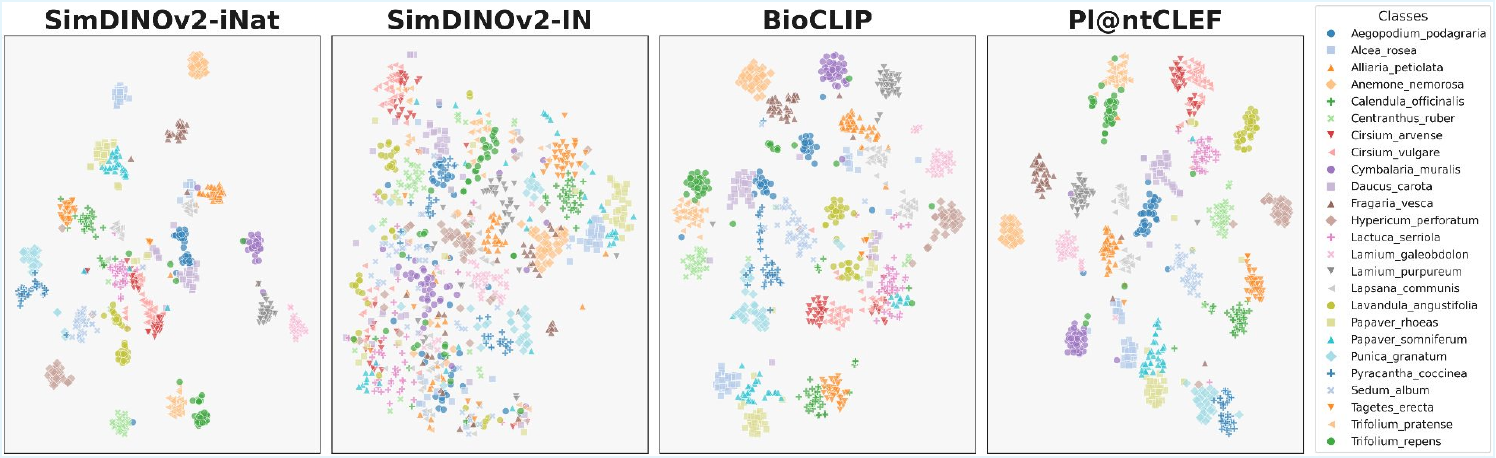}
    \caption{t-SNE visualization of feature embeddings on the Pl@ntNet dataset for different pretrained models. SimDINOv2 pretrained on ImageNet produces dispersed and overlapping class clusters, indicating poor separability of plant species. In contrast, SimDINOv2 pretrained on iNaturalist Plantae forms compact and well-separated clusters, comparable to supervised approaches BioCLIP and Pl@ntCLEF. This highlights the importance of domain-specific pretraining for fine-grained plant representation learning.}
  \label{fig:tsne}
\end{figure}

\paragraph{Feature space visualization.}
We examine the geometry of the learned representations using t-SNE projections on the PlantNet dataset (Figure~\ref{fig:tsne}). Representations obtained from ImageNet pretraining exhibit weak class structure, with samples from different species intermingled in the embedding space. In contrast, domain-specific pretraining on iNaturalist Plantae with adapted augmentations results in clearly structured clusters, where samples from the same species are grouped more coherently and separated from others. Interestingly, this organization is comparable to that observed for supervised models such as BioCLIP and Pl@ntCLEF. These results suggest that the performance gains observed in few-shot classification stem from a better alignment of the feature space with underlying semantic classes.

\paragraph{Discussion.}
These results demonstrate that domain-specific self-supervised pretraining is key to learning discriminative plant representations. Despite using no labels, our approach matches or exceeds supervised methods, highlighting the potential of SSL for scalable plant recognition.

\section{Conclusion}

In this work, we investigated self-supervised learning for plant image representation learning, a fine-grained visual recognition problem where subtle visual cues are critical for discrimination. We showed that standard augmentation strategies commonly used in SSL pipelines are not well suited to this domain, as they tend to remove or distort key discriminative features. Instead, we demonstrated that carefully designed, plant-adapted augmentations,  such as posterization and affine transformations, are more suitable for learning meaningful and discriminative representations. 

Building on these insights, we applied SimDINOv2 to plant images and showed that domain-specific pretraining on the iNaturalist 2021 Plantae subset leads to substantial improvements over generic ImageNet pretraining. Our approach achieves competitive and often superior performance compared to strong supervised baselines in few-shot settings, highlighting the effectiveness of SSL when combined with appropriate data and augmentation design. 

Overall, our results show that self-supervised learning in fine-grained domains depends on both augmentation strategies and dataset relevance, as confirmed both quantitatively through few-shot performance and qualitatively through the structure of the learned feature space. Future work should focus on scaling with larger and more diverse plant datasets to enable annotation-free plant recognition for biodiversity monitoring.

\subsubsection{Acknowledgements} This project was funded by the French National Research Agency (ANR) through the grant Pl@ntAgroEco 22-PEAE-0009. This work was granted access to the HPC resources of IDRIS under the allocation 2025-A0191011389 made by GENCI.

%
%
%
\bibliographystyle{splncs04}
\bibliography{mybibliography}

\end{document}